\documentclass[conference]{IEEEtran}

\usepackage{amsmath}
\usepackage{graphicx}
\usepackage[colorinlistoftodos]{todonotes}

\usepackage{makeidx}

\usepackage[misc]{ifsym}

\usepackage{amsmath,amsfonts,amssymb,ae,aecompl}

\usepackage{multirow}

\usepackage{bm}
\usepackage{algorithm}

\usepackage{verbatim}

\usepackage[noend]{algpseudocode}

\usepackage{url}

\usepackage{graphicx}
\graphicspath{{Figures/}}

\usepackage{xcolor}

\usepackage{amssymb}
\usepackage{graphicx}

\usepackage{balance}  
\usepackage{graphics} 
\usepackage{times}    
\usepackage{url}      

\usepackage{array,multirow}
\usepackage{longtable}
\usepackage{tikz}
\usetikzlibrary{positioning}
\usepackage{booktabs,subcaption}
\usepackage{todonotes}
\usepackage{algorithm}
\usepackage{algpseudocode}
\usepackage{amsmath}

\usepackage{algorithm, mathtools}
\usepackage{fancyhdr}
\usepackage{amsmath}
\usepackage{amssymb}
\usepackage{mathrsfs}

\newcommand{\etal}{\textit{et al.}}

\newcommand{\gc}[1]{{\textcolor{green}{#1}}}

\begin{document}

\title{The N-Tuple Bandit Evolutionary Algorithm \\ for Game Agent Optimisation}

\author{\IEEEauthorblockN{Simon M Lucas, Jialin Liu and Diego Perez-Liebana}
\IEEEauthorblockA{School of Electrical Engineering and Computer Science\\
Queen Mary University of London\\
London, UK\\
Email: \{simon.lucas, jialin.liu, diego.perez\}@qmul.ac.uk}
}


\maketitle

\begin{abstract}
This paper describes the N-Tuple Bandit Evolutionary Algorithm (NTBEA), an optimisation algorithm developed for noisy and expensive discrete (combinatorial) optimisation problems. The algorithm is applied to two game-based hyper-parameter optimisation problems. The N-Tuple system directly models the statistics, approximating the fitness and number of evaluations of each modelled combination of parameters. The model is simple, efficient and informative. Results show that the NTBEA significantly outperforms grid search and an estimation of distribution algorithm.
\end{abstract}



\section{Introduction}\label{sec:intro}


This paper describes the N-Tuple Bandit Evolutionary Algorithm (NTBEA) and its application to optimising the parameters of a rolling horizon evolution game-playing agent.  The NTBEA combines evolutionary search with Multi-Armed Bandit algorithms (MABs) in order to provide an algorithm which is robust to noise, has an explicit way to balance the trade-off between exploration and exploitation, and provides a statistical model of the fitness landscape as an additional output.

The applications of this type of algorithm are numerous. In our research we have already applied it successfully to hyper-parameter optimisation~\cite{sironi2018self} and automated game tuning~\cite{kunanusont2017ntuple}.
Furthermore, if the inherent fitness landscape is flat, then the exploration term provides a means for performing novelty search~\cite{Lehman2011Novelty}.

\subsection{Estimation of Distribution Algorithms}
Estimation of Distribution Algorithms (EDAs)~\cite{muhlenbein1996recombination,larranaga2001estimation,gonzalez2002mathematical,lozano2006towards,hauschild2011introduction} are a powerful class of Evolutionary Algorithms (EAs).
Instead of using the mutation and crossover operators to generate offspring from the fittest parents,  the population is sampled from an estimated probability distribution of selected individuals, which is updated iteratively using the fitness evaluations of individuals.
In addition to potentially making the search more efficient and robust to noise, EDAs also have the benefit of learning a model which offers greater insight into the nature of the problem. The model may provide information regarding key parameter combinations that tend to lead to good or bad solutions.

The NTBEA models not only the fitness of points in the search space, but also
estimates how much each point has been visited.  This enables explicit modelling of exploration, which has two distinct benefits. First, it enables a principled way to avoid becoming trapped in local optima, since the search can be 
diverted to less well explored areas. Second, it provides a natural way to perform a type of novelty search. This is especially important when dealing with problems that involve a flat reward landscape~\cite{decock2013noisy}.


\subsection{Combinatorial Multi-Armed Bandits}
Rolet and Teytaud~\cite{rolet2010bandit} modeled the sub-domains of the search space as arms of a bandit and allocated the computational resource dynamically during the evaluation step of an evolution strategy for continuous noisy optimization.
When searching in a discrete space, the NTBEA has significant similarity to the Combinatorial Multi-Armed Bandits (CMABs), formalised by~Gai~\etal~\cite{gai2010learning} and Chen~\etal~\cite{chen2013combinatorial}. Gai~\etal~applied CMABs to a channel allocation problem in mobile communications, and showed how considering pair-wise interactions led to better performance than a naive univariate model, but their model was particular to resource allocation problems.  
The recent work of Onta\~n\'on \cite{ontanon2017combinatorial} on CMABs is the closest to this paper.
Onta\~n\'on used CMABs for multi-agent control in the microRTS game, a relatively simple but highly challenging real-time strategy (RTS) game.
Onta\~n\'on's CMABs model each dimension of the search space as an individual local MAB, in which each arm is a legal value of the corresponding dimension, together with a global MAB with all the legal value combinations as arms. These value combinations are sometimes called macro-arms.

\subsection{N-Tuple Systems}
In this paper we use an N-Tuple system to model the fitness landscape, and combine it with an evolutionary search algorithm.  
N-Tuple systems were first developed and applied to pattern recognition in the late 1950s \cite{NTuple1959}, and among many other applications have also been applied to model value functions in games \cite{NTupleOthello,NTuplePreference}.
The standard way of using an N-Tuple system is to model a high-dimensional feature space by considering a set of lower-dimensional projections of that space, defined by the set of N-Tuples.  
Each N-Tuple forms a weak model of the space, but combining them provides a good model.
There are some novel aspects of the N-Tuple systems used in this paper that will be described later in Section \ref{sec:ntbea}.

The notion of successively improving an individual or a population via variation operators and fitness evaluations makes the NTBEA distinct from the standard CMAB methods. 
A CMAB operates via a pure bandit-based sampling procedure, whereas the approach described in this paper uses an evolutionary algorithm (EA) to perform the population generation and variation, and a bandit landscape fitness model (for brevity called just the model or the model space) to perform the selection.  In this paper an N-Tuple system is used as the model, but other choices of
model would also be possible.  N-Tuple systems are a good choice due
to their speed, simplicity, interpretability and reasonable accuracy.
Essentially, the EA searches the model space to improve the sample efficiency in the problem domain.  

In this paper, a variation of the $(1,\lambda)$-EA is used, in which the current best individual is selected as the parent only to generate the next population of size $\lambda$ and is not included in the new population.

The instance of the algorithm described in this paper is similar to the N-Tuple Bandit Evolutionary Algorithm introduced by 
Kunanusont~\etal~\cite{kunanusont2017ntuple}, extended from the Bandit-based Random Mutation Hill Climber~\cite{liu2017bandit}.  
Kunanusont~\etal~\cite{kunanusont2017ntuple} applied the NTBEA to automatic game parameter tuning of a two-player adversarial video game, a very noisy problem, as some stochastic AI agents were used to evaluate the evolved stochastic games.
The NTBEA significantly outperformed two evolutionary algorithms, a simple Random Mutation Hill Climber and a univariate Bandit EA~\cite{kunanusont2017ntuple}.
The NTBEA was also applied within the General Video Game AI (GVGAI~\cite{perez2016general}) framework for evolving game rules and parameters, and has successfully evolved variations of the game \emph{Aliens} to favour either a short-term or a long-term planning agent~\cite{tutorial17cig}.


In this paper, we formalise the NTBEA, and present its applications for hyper-parameter optimisation. The NTBEA makes use of the statistics of the previously evaluated solutions and balances the trade-off between exploring a new (or less evaluated) solution and exploiting the best one found so far.
The NTBEA is particularly useful when the evaluation function is expensive or noisy. For example, in the context of automatic game design, a game needs to be played several times to approximate an accurate win rate if the game or the agents are stochastic, while the execution time of a single game may take a significant amount of time or it is difficult to perform an evaluation.  
The notion of \emph{expensive} in this context is clearly subjective, but really means any problem where the evaluation time is a significant issue in the opinion of the algorithm user.

Using N-Tuples to approximate fitness statistics and visit counts can be related to locally sensitive hashing (LSH), and it is worth noting how the N-Tuple system described in this paper could also be used for enhancing exploration in video games, similar to the way LSH in the form of Context Tree Switching (CTS) was applied to boost performance on the challenging game of Montezuma's Revenge~\cite{VisitCountExplorationNIPS2016}.

\subsection{Hyper-Parameter Optimisation}

The task of tuning the parameters of a game-playing agent can be viewed
as a Hyper-Parameter Optimisation problem.
Bergstra and Bengio~\cite{bergstra2012random} state how the most popular methods are Grid Search and manual search, but show how competitive the performance of random search is for these problems.  More recently the Hyperband paper~\cite{li2016hyperband}
shows how recent approaches such as Sequential Model-based Algorithm Configuration (SMAC) \cite{SMAC-LION-2011}
and Tree-structured Parzen Estimator (TPE;~\cite{bergstra2011algorithms}) in
some cases only perform similarly to random search. They outperformed random search in a direct rank-based comparison, but only by a small margin. When the random search was repeated twice and the best solution was picked, random search performed best \cite{li2016hyperband} (albeit not quite a fair comparison, but one that illustrates the small margin of improvement).

The NTBEA approach has a similar overall algorithmic framework to SMAC \cite{SMAC-LION-2011},
in the way a model is iteratively updated and used to decide which point in the search
space to sample next.  Where SMAC uses random forests for the model, the NTBEA uses N-Tuple
systems.  There are also significant differences in the approaches; SMAC only estimates the value
of the objective function for each parameter configuration, whereas NTBEA also estimates the 
visit count, which is essential for the bandit-based exploration pressure.  There are also some
conceptual differences, with SMAC running $T$ trials of each sample configuration (via its \emph{INTENSIFY}
procedure) while by default NTBEA runs a single evaluation each time, in order to make maximum
use of the bandit model.

Other aspects of hyper-parameter optimisation in general
involve early stopping of unpromising solutions,
and also sharing of the system parameters between competing
solutions: both of these are used to great effect by Jadeberg 
\emph{et al} \cite{PopTrainNeuralNets}.  
Here we only consider when each solution is fully evaluated at least once,
as in a complete game is played, though actually games could be abandoned
very early when one player is obviously weak.
For now we note that there are many approaches to hyper-parameter
optimisation, but that manual and grid search are in strong use,
and that random search performs surprisingly well.

The rest of this paper is structured as follows.
Section~\ref{sec:ntbea} introduces the overall structure of the NTBEA.
The test problem is described in Section~\ref{sec:problem}, then results are illustrated and discussed in Section~\ref{sec:res}.
Section~\ref{sec:conc} concludes and lists some future work.

\section{The N-Tuple Bandit Evolutionary Algorithm}\label{sec:ntbea}

\begin{figure}[hbtp]
\centering
\includegraphics[width=1\linewidth]{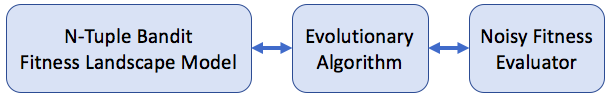}
\caption{\label{fig:bleda}Key components of the N-Tuple Bandit Evolutionary Algorithm (NTBEA). 
A model of the fitness landscape is built and updates iteratively using the evaluations of solutions.
The search for candidate solutions is performed in the model space, i.e., the N-Tuple Bandit Fitness Landscape model, where candidate solutions can be evaluated quickly (N-Tuple systems are well known for their speed~\cite{rohwer1996theoretical}, and our experimental tests support this). 
This is then sampled in the real (relatively expensive) problem search space, where the fitness value is expected to be noisy. The model is then updated with the fitness value of this solution point. The process repeats until some termination condition is met.}
\end{figure}

In this section, we describe the system architecture of the algorithm and then explain each part in more detail.
Figure~\ref{fig:bleda} illustrates the three key components of the NTBEA, a bandit landscape model, an evolutionary algorithm and a noisy evaluator, i.e., a fitness function corrupted by noise.
We assume that the execution time of querying the bandit landscape model is negligible compared to evaluating a candidate solution on the target problem.

The algorithm works as follows. The search begins by sampling a single solution point uniformly at random in the search space.
This search point is referred to as the \emph{current point}.
The fitness of the current point is evaluated once, in the problem domain, using the noisy evaluator. In theory, no resampling is needed, even for noisy problems, since the UCB module in the algorithm aims to take care of re-evaluating an identical point if needed. 
Note that the algorithm also works for noise-free
problems, though we focus on noisy problems in this paper.

The current point is then stored in the bandit fitness landscape model (referred to as the \emph{model} from now on), together with its fitness value. The model is then searched within the neighborhood of the current point. The neighborhood is defined using the number of neighbors and the proximity distribution to the current point, which is controlled by a mutation operator. The solution point in the neighborhood with the highest estimated Upper Confidence Bounds value (UCB; defined later in Section~\ref{sec:ucb}) is then selected as the new current point. The process iterates until the evaluation budget is used up, or some other termination condition is met.  The algorithm is described in Algorithm~\ref{algo:ntbea}.

Note that this description outlines the simplest
approach where only a single solution is held as the current focus of the search, similar to using only one parent. Population-based versions are also possible, and would be more appropriate for parallel hardware.

\subsection{Estimating UCB Values: an $N$-Tuple Approach}\label{sec:ucb}

A key part of the algorithm is the value function used to sample in a large search space.  In this context, \emph{large} means that the size of the search space is larger than the number of fitness evaluations allowed, thus it is impossible to properly evaluate (proper evaluation may involve resampling due to noise) each of the possible solutions points.
Instead, we need to model the relationship between points in the search space and sample accordingly.


The algorithm UCB1, a simple multi-armed bandit algorithm, is introduced by Auer~\etal~\cite{auer2002finite}. The UCB value of any arm $i$ is defined as:
\begin{equation}\label{eq:ucb}
\mbox{UCB}_i = \hat{X_i} + k \sqrt{\frac{ \ln n }{ n_i + \epsilon}}.
\end{equation}
The empirical mean reward for playing arm $i$ is $\hat{X_i}$: this is the exploitation term. The right hand term controls the exploration, where $n$ is the total number of times this bandit has been played, and $n_i$ is the number of times the arm $i$ has been played. The term $k$ is called the exploration factor: higher values of $k$ lead to explorative search, low values lead to more greedy or exploitative search. Each dimension of the search space is modelled as an independent MAB, with an arm for each possible value. The $\epsilon$ value is used to control whether each arm should be pulled at least once. In the standard UCB formula, $\epsilon$ is set to zero ensuring that each arm is pulled once in turn, but for our purposes this would be impractical, as the macro-arm consisting of the entire $N$-Tuple would force an exhaustive exploration of the search space.

Additionally, we also model combinations of arms as \emph{super-bandits}.  For example, in a $d$-dimensional search space where each dimension has $n$ possible values, the 10-wide super-bandit would have $n^d$ arms.

We modify this to be an aggregate over all the $N$-Tuples in the $N$-Tuple System model.
Let $N$ be the $N$-Tuple indexing function such that $N_j(x)$ indexes the $j^{th}$ bandit for search space point $x$.
Initially we compute the aggregate UCB value for solution point $x$ as an unweighted arithmetic average, as defined in (\ref{eq:ucbagg}):
\begin{equation}\label{eq:ucbagg}
v_{UCB}(x) = \frac{1}{m}\sum_{j=1}^{m} \mbox{UCB}_{N_j(x)},
\end{equation}
where $m$ denotes the total number of bandits in the system.

More sophisticated algorithms are possible, for example by combining the individual outputs using a Bayesian method, but for simplicity and proof of concept we begin by using the arithmetic average.

The N-Tuple systems have ideal properties for use as fitness landscape models, in that they offer an extremely fast one-shot training and good accuracy.
They are ideally suited to modelling discrete spaces, but can also be applied to continuous spaces with some degree of compromise.  In this paper we are dealing with discrete search spaces, so they are already a good fit. 

The concept is as follows. Given a $d$-dimensional search space, we sub-sample its dimensions with a number of $N$-tuples. The value of $N$ ranges from $1$ up to $d$, though may miss out values in between. If all the tuples were considered, then the total number of bandits is $2^d$.

In the standard N-Tuple systems each entry in the look-up-table stores a single value for each class, normally related to the probability of that index occurring given that class, or for game position value function approximation, the value of that index occurring.

In our model, however, each N-Tuple has a look-up table (LUT) that stores statistical summaries of the values it encounters; the basic numbers stored are the number of samples, the sum of the fitness of samples, and the sum of the square of the fitness of the samples. This enables the mean, the standard deviation and the standard error to be calculated for each entry in the table. This provides all we need for (\ref{eq:ucb}) and (\ref{eq:ucbagg}), and beyond: calculating the mean, standard deviation or standard error for each parameter combination being modelled provides useful insight into the nature of the system under optimisation.


\subsection{Algorithm}
\begin{algorithm}[!t]
\caption{\label{algo:ntbea}The N-Tuple Bandit Evolutionary Algorithm.}
\begin{algorithmic}[1]
\Require{$\mathbb{S}$: search space}
\Require{$fitness$: noisy solution evaluator}
\Require{$n \in N^+$: number of neighbors}
\Require{$p \in (0,1)$: mutation probability}
\Require{$flipOnce \in \{true, false\}$: flip at least once or not}
\Require{$nbEvals$: total number of evaluations allowed}
\State{$t = 0$}\Comment{Counter for fitness evaluations}
\State{$Model \leftarrow \varnothing$} \Comment{Initialise the fitness landscape model}
\State{$current \leftarrow$ a random point $\in \mathbb{S}$}
\While{ $ t < nbEvals$}
\State{$value \leftarrow fitness(current)$} \label{lst:line:fitness}
\State{add $<current,value>$ to $LModel$} \label{lst:line:lmodel}
\State{\scriptsize $Population \leftarrow$ ~\Call{Neighbors}{$Model,current,n,p,flipOnce$}} \label{lst:line:neigh}
\State{$current \leftarrow  \arg\max_{x \in Population} v_{UCB}(x)$} \label{lst:line:current}
\State{$t \leftarrow t+1$}
\EndWhile
\State{\Return{ $LModel$}}
\Statex
\Function{Neighbors}{$model,x,n,p,flipOnce$}
\State{$Population \leftarrow \varnothing$} \Comment{Initialise empty set}
\State{$d \leftarrow |x|$} \Comment{Get the dimension}
\For{$k \in \{1,\dots,n\}$}
	\State{$neighbor \leftarrow x$}
    \State{$i \leftarrow 0$}
	\If{$flipOnce$}
        \State{$i\leftarrow$ randomly selected from $\{1,2,\dots,d\}$} 
    \EndIf
	\For{$j \in \{1,\dots,d\}$}
    \If{$i==j$}
		\State{Randomly mutate value of $neighbor_j$}
	\Else
		\If{$RAND < p$} 
          	\State{Randomly mutate value of $neighbor_j$}
        \EndIf
	\EndIf
\EndFor
    \State{Add $neighbor$ to $Population$}
\EndFor
\State \Return{$Population$}
\EndFunction
\end{algorithmic}
\end{algorithm}
The NTBEA algorithm is outlined in Algorithm~\ref{algo:ntbea} and operates as follows. It begins by choosing a random solution point in the search space, which is called the current point. 
Since we are dealing with discrete search spaces, each point is represented as a vector of integers, where each element of the vector is an index to the currently selected value in that dimension. The actual value chosen may be of any type, common types are integer, double and boolean.

The following steps are then repeated until the fitness evaluation budget has been exhausted:
\begin{itemize}
\item It makes a (noisy) fitness evaluation of the current point and stores it in the N-Tuple Fitness Landscape Model as the value for that solution point (lines~\ref{lst:line:fitness} and~\ref{lst:line:lmodel} in Algorithm~\ref{algo:ntbea}).
\item Using a mutation operator to generate a set of unique neighbors of the current solution, as the population of current iteration (line~\ref{lst:line:neigh}). 
\item  Using the fitness landscape model, the algorithm calculates the estimated UCB value of each solution by (\ref{eq:ucb}) and (\ref{eq:ucbagg}). Then, it sets the current solution as the one in the population (neighbors from the previous step) with the highest estimated UCB value  (line~\ref{lst:line:current}).
\end{itemize}
When the fitness evaluation budget has been exhausted, the method searches and recommends a neighbor of all of the evaluated solutions, in which each dimension is set to the value with maximal approximate value defined in (\ref{eq:ucbagg}). 





\subsection{Illustrative Example}

Consider a $5$-dimensional space, modeled using five $1$-tuples and one $5$-tuple. Suppose we now enter the four vectors (three unique) in the search space together with their associated fitness values as given in Table \ref{tab:entries}.

The first 1-tuple will have a LUT that has 2 entries, with $LUT[0****]$ having a mean of 1, and $LUT[1****]$ having a mean of $\frac23$. The single 5-tuple will have three non-empty entries, with $LUT[12340]$ having a mean of 0.5, and $LUT[11111]$ and $LUT[00110]$ both having means of 1. Note that the object at each index is a statistical summary object as described above that does not directly store the mean but can calculate it; we describe the mean value to best illustrate the operation, and because it feeds directly into (\ref{eq:ucb}) and (\ref{eq:ucbagg}).
An example of some of the statistics that are stored in the system or can be calculated directly is shown in Table \ref{tab:stat}.

\begin{table}[!t]
\centering
\caption{\label{tab:example}Illustrative example, in which only $1$-tuples and $d$-tuple are considered ($d=5$).}
\begin{subtable}{1\linewidth}
\centering
\caption{\label{tab:entries}Evaluated solutions (entries) and corresponding fitness values.}
\begin{tabular}{cc}
\hline
Solution & fitness\\
\hline
 $[1,2,3,4,0]$ & 1\\
 $[1,1,1,1,1]$ & 1\\
 $[0,0,1,1,0]$ & 1\\
 $[1,2,3,4,0]$ & 0\\
\hline
\end{tabular}
\end{subtable}
\begin{subtable}{1\linewidth}
\centering
\caption{\label{tab:stat}Some of the statistics that are stored in the system or can be calculated directly.}
\begin{tabular}{cccc}
\hline
$N$-tuple & Pattern & Mean & Nb. of eval.\\
\hline
\multirow{11}{*}{1-tuple} & $[0,*,*,*,*]$	& 1	& 1\\
& $[1,*,*,*,*]$	& $\frac23$	& 3\\
& $[*,0,*,*,*]$	& 1	& 1\\
& $[*,1,*,*,*]$	& 1	& 1\\
& $[*,2,*,*,*]$	& $\frac12$	& 2\\
& $[*,*,1,*,*]$	& 1	& 2\\
& $[*,*,3,*,*]$	& $\frac12$	& 2\\
& $[*,*,*,1,*]$	& 1	& 2\\
& $[*,*,*,4,*]$	& $\frac12$	& 2\\
& $[*,*,*,*,0]$	& $\frac23$	& 3\\
& $[*,*,*,*,1]$	& 1	& 1\\
\hline
\multirow{3}{*}{5-tuple} & $[0,0,1,1,0]$	& 1	& 1\\
& $[1,1,1,1,1]$	& 1	& 1\\
& $[1,2,3,4,0]$	& $\frac12$	& 2\\
\hline
\end{tabular}
\end{subtable}

\end{table}

In this example only the empirical average and the number of evaluations is output for each index of each $N$-tuple, though as mentioned previously the standard deviation and standard error are also available. For each $N$-tuple, only the non-null table entries are shown, i.e., ones in which the index (can also be thought of as a pattern) occurs at least once.







\section{Noisy Optimisation of Game Agent Parameters}\label{sec:problem}

This section describes an application to optimising the parameters of a rolling horizon evolutionary game-playing agent. The agent is optimised to get as high average score on the game as possible. This is a noisy optimisation problem with two distinct sources of noise: (i) the game itself is stochastic; (ii) the optimised agent follows a stochastic policy, such that given the same game state, it may play differently if simulating more than once.

The behavior of the agent is controlled by a number of parameters, some of which may have a critical effect on the performance of the agent, and where the combination of parameter values is important.

The problem addressed is similar to the hyper-parameter optimisation problem that is  topical in machine learning; it has been shown many times that optimising a known architecture can produce state of the art results (e.g., \cite{bergstra2011algorithms,bergstra2012random}).

\subsection{Rolling Horizon Evolutionary Agent}\label{sec:rhea}

We aim to optimise the parameters of a rolling horizon evolutionary agent. 
Rolling Horizon Evolutionary Algorithms (RHEAs) are called \emph{rolling horizon} as each of the individuals in the population is an action sequence of a fixed planning (time) horizon, $h_p$, thus RHEAs plan ahead $h_p$ actions. On each subsequent game step, the horizon \textit{rolls} one step further within the $h_p$ window.
Every individual is evaluated by evaluating the state after simulating $h_p$ actions in its corresponding sequence or earlier, if a termination state is reached before performing all the $h_p$ actions. 
Then, only the first action of the best individual is applied. 
This procedure repeats with the updated state.

The \emph{rolling horizon} technique can be integrated to different evolutionary algorithms such as Genetic Algorithms and Coevolutionary Algorithms~\cite{liu2016rolling}. In this work, we combined a simple $(1+1)$-EA with \emph{rolling horizon}. This agent will be referred to as $RHEA$ in the rest of the paper.

The performance of RHEAs can be boosted in many ways. For instance, recently, RHEAs have been applied to General Video Game Playing, and proved that the population size and sequence length (i.e., planning horizon) had significant effect on the performance of the algorithms~\cite{gaina2017analysis,gaina2017population,gaina2017rolling}.
Some of the key parameters are
\begin{itemize}
\item \emph{sequence length}: planning horizon;
\item \emph{shift buffer} enabled or not: if disabled, at any timestep $t+1$, the initial population is reset to random, otherwise, each of the individuals from the population at timestep $t$ (previous optimisation procedure) shifts its action sequence forward and fills the last position by a random action;
\item \emph{resampling number}: in the stochastic case, how many time an individual is re-evaluated (equals to $1$ if no re-evaluation is allowed).
\item \emph{mutation probability}: how likely a mutation occurs at every dimension;
\item \emph{flip at least one bit}: indicates if at least one mutation should occur at each time.
\end{itemize}

\def\useless{Note that the tick budget is used to derive the number of evaluations of
the rolling horizon agent each time: this is the number of game ticks allowed
for each decision made by the game-playing agent.}

\subsection{Test Problems}
We optimise the rolling horizon evolutionary agent to play two games: simplified Java implementations of Asteroids and Planet Wars.
In these test cases, the $RHEA$ agent model sequences of actions in next game ticks as its individuals, and evolve the individuals over time aiming at maximising some fitness value.

\subsubsection{Asteroids}\label{sec:asteroids}
Asteroids, released to great acclaim in 1979 is one of the classic video games of all time, and Atari's most profitable\footnote{\url{https://en.wikipedia.org/wiki/Asteroids_(video_game)}}.
The original game was implemented in special vector graphics hardware, featured memorable sound effects and smooth physics-based movements.
The challenge for players is to avoid being hit by asteroids, while aiming at them accurately and to also hit the flying saucers before they hit the player's spaceship.  
There are three sizes of asteroids: large, medium and small.  
Each screen starts with a number of large rocks: as each one is hit either by the player, or by a missile fired by an enemy flying saucer, it splits in to a smaller one: large rocks split into two medium ones, medium into two small ones, small ones disappear.  
There is also a score progression, with the score per rock increasing as the size decreases.  

Strong play requires good perceptual and motor skills for avoiding collisions and shooting accurately at the targets.
There are also some interesting strategies which may easily elude novice players.  
One of them is to control the number of asteroids on the screen by shooting one large one at a time, then picking off a single medium rock and each small one it gives rise to, before breaking another large rock.

The largest score for any item is the $1,000$ points awarded for shooting the small flying saucer.  
These fire at the player with deadly accuracy, so it is important to shoot them quickly as the missiles they fire are fast moving and tricky to avoid.  
Many strong players will shoot all but one remaining asteroid (a small one) and then lie in wait for flying saucers at the edge of the screen, firing at them immediately as soon as they appear.  
There is some devil in the detail here, regarding the best place to wait and the best angle to fire at (the screen wraps around both vertically and horizontally, so from a single position it is possible to shoot at both sides).

For our experiments we used a Java implementation of the game where we can simulate the game $10,000$ times faster than the real game.
Although using the ALE or MAME version of the game would have been possible, there are some distinct advantages to having our own implementation:
\begin{itemize}
\item It runs much faster than the ALE or MAME versions.
\item Having access to the source code makes it straightforward to vary details of the game in order to test various hypotheses.  For example, we can test the effects of penalising each missile fired, or giving the large and medium rocks zero value to test the long-term planning ability of the agents under test. An example is the use of the two-player version of this implementation in automatic game parameter tuning~\cite{liu2017evolving}.
\end{itemize}
In this simplified version, there are at each time twelve legal actions at each game tick, being a combination of the three steer actions (\emph{LEFT}, \emph{CENTER}, \emph{RIGHT}), and whether the ship is thrusting ( \emph{THRUST}) and / or firing (\emph{FIRE}).  Unlike the original game there is no \emph{HYPERSPACE} action. The player starts with $0$ as game score and $3$ lives, and an additional life every $10,000$ points.
The player gets $200$, $100$ or $50$ points every time it hit an large, medium or small asteroid, respectively, and loses $10$ points for every missile fired.
Losing a life has a penalty of $-200$ points. The game stops when all lives are lost, or $1,000$ game ticks have elapsed.  Currently there are no flying saucers, but we plan to add this feature for the next set of experiments.

Figure \ref{fig:Asteroids} gives a screenshot of the game screen. 
Pink lines illustrate the simulations of the $RHEA$ agent which controls the spaceship; these are shown for illustrative purposes only, and ignore the
fact that each rollout involves other changes to the game state such
as firing missiles and the rocks moving and splitting.

\subsubsection{Planet Wars}
Planet Wars is a simple but challenging Real Time Strategy game
that was run as a highly successful Google Game AI Challenge in 2011
by the University of Waterloo in Canada.
For the work in this paper we implemented a simpler version of
the game to make it faster to allow rapid running of experiments
while retaining some challenging aspects of the original game.
The game runs at more than $10$ million game ticks per second. 

The aim game of Planet Wars is to occupy all the planets, where
each planet is occupied either by player 1, player 2, or by a neutral.
Each planet has a number of ships belonging to the owner of the planet.  
To invade a planet a player sends a number of ships to the planet to be invaded
from the planet it owns.
For our experiments we used an implementation with the following rules:

\begin{itemize}
\item No neutral planets: the ships on each planet are either owned by player 1 or player 2.
\item At each game tick, a player moves by shifting ships to or from a player's ship buffer, or by moving it's current \emph{planet of focus}.
\item When a player transfers ships it is always between it's buffer and the current planet of focus.
\item At each game tick the score for each player is the sum of all the ships on each planet it owns, plus the ships stored in it's buffer.  We have two versions of the game: the easiest for the planning agents, and the one used in this paper also adds in the ships in a player's buffer to it's score, a more deceptive version of the game does not include this.  
\end{itemize}

For research purposes, the advantage of the modified action space is that
it makes it compatible with the General Video Game AI framework, giving
direct access to a large number of game-playing agents for comparison 
purposes.  

\section{Applications and Results}\label{sec:res}
We aim to optimise the parameters of the $RHEA$ agent (described in Section \ref{sec:rhea}) with NTBEA and two baseline algorithms, Grid Search and a multi-valued version of the Sliding Window compact Genetic Algorithm (SWcGA), an EDA proposed by Lucas~\etal~\cite{lucas2017efficient} recently.\footnote{The original Compact Genetic Algorithm (cGA) and the sliding window version described by Lucas~\etal~\cite{lucas2017efficient} only handled binary strings.  The
version used in this paper handles integer strings and can use absolute or
relative fitness measures; the version used here used absolute fitness measures.}
The NTBEA used in the experiments takes into account all the $d$ 1-tuples, $\frac{d(d-1)}{2}$ 2-tuples and the only $d$-tuple.

\subsection{Experimental Setting}

\subsubsection{Fitness Function}
In each case the fitness function is based on the outcome of
a single game, which leads to noisy fitness functions.
\paragraph{Asteroids}
The evaluation function is the game score when the game terminates, the higher the better.
The calculation of the game score is explained in Section \ref{sec:asteroids}.
The value of $k$ should be set relative to the score distribution;
for Asteroids $k$ was set to $5,000$.

\paragraph{Planet Wars}
The evaluation function is based on playing a single game, with a value of $+1$ if the agent wins or $-1$ for a loss, and $0$ for a draw (which is very unlikely to occur).  This gives an extremely noisy fitness function.
Note that an optimiser can choose to resample a particular point in the search space, i.e., playing multiple games using the agent with an identical parameter setting, in order to get an estimate of win rate. However, with a fixed small evaluation budget, the optimiser will run fewer iterations compared to sampling exactly one.  

The NTBEA optimises the parameters of a $RHEA$ agent
controlled as player 1, versus a fixed-parameter version of the $RHEA$ with well-chosen parameters based on the authors' experience playing as player 2.  The game is symmetric and the planning budget for both players was set to $2,000$ game ticks.
For Planet Wars, the value of $k$ in the NTBEA  was set to 1.0.

\subsubsection{Search Space}
The $RHEA$ agent is characterised by the parameters shown in Table \ref{tab:space}, where each one is given a type and legal values. The parameters have been detailed previously in Section \ref{sec:rhea}. The mutation probability for a $d$-dimensional problem is calculated by $nbMutatedPoints/d$.

\begin{table} [!t]
\centering
\caption{\label{tab:space}Search space of the parameter settings.}
\begin{tabular}{cccc}
\hline
\multicolumn{2}{c}{Variable} & Type & Legal values \\
\hline
\multirow{2}{*}{$sequenceLength$} & Asteroids & Integer & $5, 10, 15, 20, 50, 100, 150$ \\
& Planet Wars & Integer & $5, 10, 15, 20, 50$ \\
\multicolumn{2}{c}{$nbMutatedPoints$} & Integer & $0.0, 1.0, 2.0, 3.0$ \\
\multicolumn{2}{c}{$flipAtLeastOneBit$} & boolean & $false, true$ \\
\multicolumn{2}{c}{$useShiftBuffer$} & boolean & $false, true$ \\
\multicolumn{2}{c}{$nbResamples$} & Integer & $1, 2, 3$\\
\hline
\end{tabular}
\end{table} 

The $RHEA$ agent is
very flexible and allows any compatible evolutionary algorithm to
be plugged in to control the evolutionary process.  For these experiments
we used a Random Mutation Hill Climber / $(1+1)$-Evolutionary Algorithm (the distinction between the two depends on the parameters chosen by the hyper-parameter optimiser).

The search space of the parameter settings is actually the search space of $RHEA$ instances, while every $RHEA$ given a parameter setting is considered as a distinct instance.
As shown in Table \ref{tab:space}, the size of the search space, in other words, total number of possible $RHEA$ instances, when playing Asteroids and Planet Wars, is $336$ and $240$, respectively.

\subsubsection{Budget}
We allowed an optimisation budget of the same value as the size of search space, thus $336$ game evaluations in the case of Asteroids and $240$ in the case of Planet Wars.
Note that a ``game evaluation'' refers to a whole game playing, which lasts at most $1,000$ game ticks. At each game tick, the playing agent $RHEA$ has a budget of $2,000$ forward model calls (simulations) to find an optimal action to play.
Given that there are $336$ and $240$ points in the search space of the two games, this means an attempted uniform Grid Search cannot even sample each point twice.
The NTBEA works by building up statistics for each value in each individual dimension, as well as tuples of values, and the full-width n-tuple. Hence the optimiser is able to make good use of the gathered information.

\subsection{Results}
Each of the three optimisation algorithms, Grid Search, SWcGA and NTBEA, are given the same budget to optimise a $RHEA$ agent on each of the games. The recommended agent instance at the end of an optimisation is validated by playing $100$ games.
NTBEA was run with a neighborhood-size of 50, and SWcGA with a sliding window of
size 50.

Table \ref{tab:res} analyses the average fitness value obtained by the $RHEA$ instances recommended by three tested algorithms over 10 optimisation trials on Asteroids and Planet War. $RHEA_{GridSearch}$, $RHEA_{SWcGA}$ and $RHEA_{NTBEA}$ denote the $RHEA$ agent instance recommended at the end of optimisation by Grid Search, SWcGA and NTBEA, respectively.
A random agent, which uniformly randomly selects an action at every game tick, is also tested for comparison.
Unsurprisingly, 
the random agent performs poorly both in games. 

\begin{table} [!t]
\centering
\caption{\label{tab:res}Average fitness value over 10 optimisation trials. Standard error is given after $\pm$. Note that in the case of Planet War, an average of $0$ refers to a win rate of $50\%$.}
\begin{tabular}{lrl}
\hline
\multicolumn{1}{c}{Agent} & \multicolumn{1}{c}{Planet Wars} & \multicolumn{1}{c}{Asteroids}\\
\hline
$Random$ & -0.9400 $\pm$ 0.3400 & 1,091 $\pm$ 88 \\
$RHEA_{GridSearch}$  & -0.53 $\pm$ 0.12 & 7,716 $\pm$ 330 \\
$RHEA_{SWcGA}$ & 0.39 $\pm$ 0.07 & 8,439 $\pm$ 110 \\
$RHEA_{NTBEA}$ & \textbf{0.54 $\pm$ 0.02} & \textbf{8,756 $\pm$ 41} \\
\hline
\end{tabular}
\end{table}

Given the same budget, the NTBEA significantly outperforms the baseline algorithms, Grid Search and SWcGA.  Of the three, Grid Search is by far the worst.
The Wilcoxon Signed-Rank Test was run for significance between $RHEA_{SWcGA}$ and $RHEA_{NTBEA}$, establishing a significance difference between the measures with $p = 0.01928$ for Planet Wars. The results for Asteroids follow a normal distribution under the Shapiro-Wilk Normality Test, and the difference between $RHEA_{SWcGA}$ and $RHEA_{NTBEA}$ is significant with a student t-test p-value of $0.040131$.

In addition to the results presented in Table~\ref{tab:res} the authors have
also made many other test runs with various evaluation budgets, both
higher and lower than the budgets used for the table, and
in all the tests made so far the NTBEA has outperformed the SWcGA,
and also outperformed grid search where applicable (grid search was not
applied to cases where the budget was smaller than the number of points in the search space).  A more thorough set of experiments, including ones with much larger search spaces is on-going work.

\begin{figure*}[hbtp]
\centering
\begin{subfigure}[t]{0.47\textwidth}
\centering
\includegraphics[height=.7\linewidth]{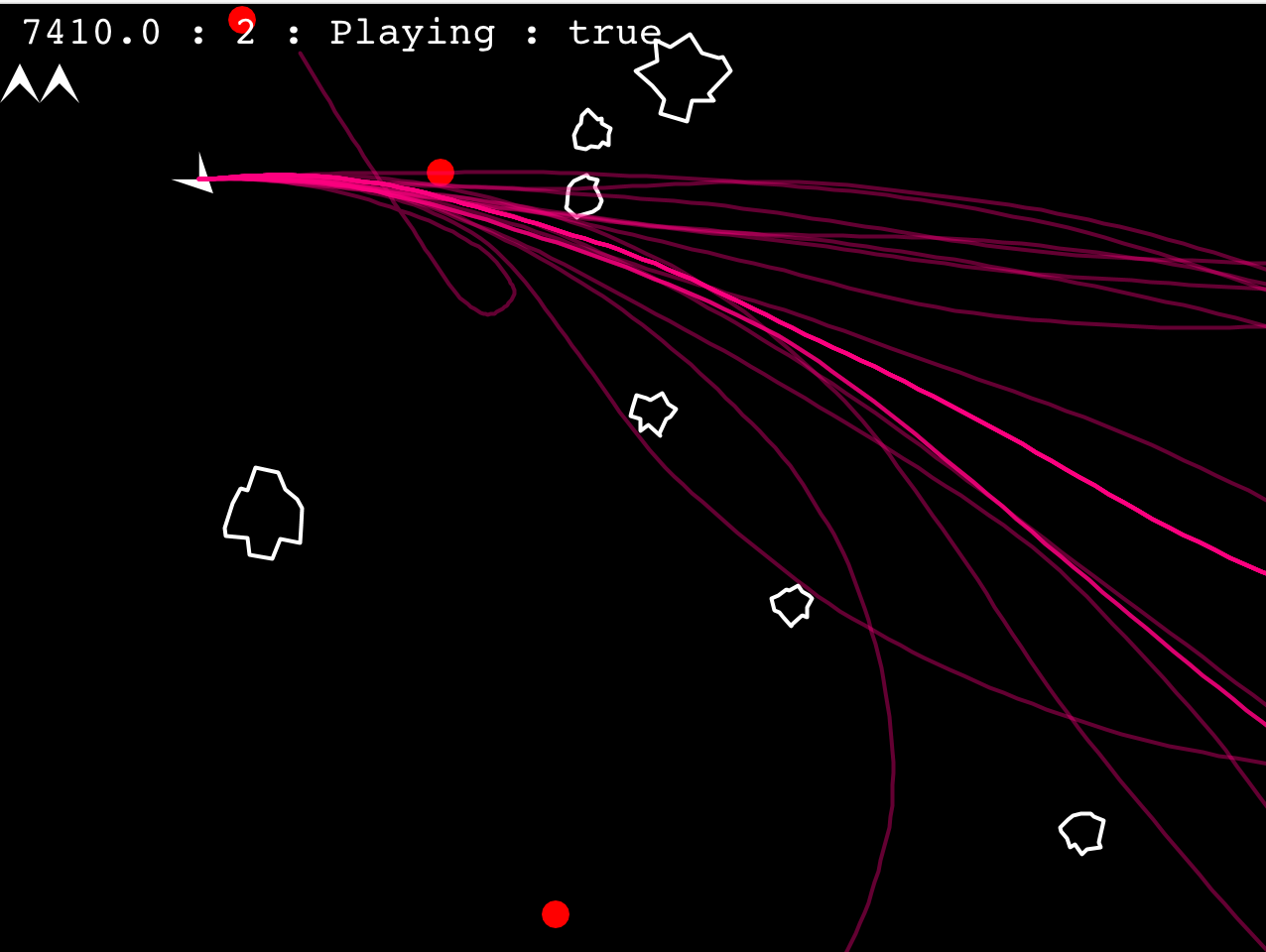}
\caption{\label{fig:Asteroids}Screenshot of the game screen of Asteroids. Polygons represent asteroids. Solid rectangle represents the spaceship, and pink lines illustrate the simulations of the rolling horizon evolutionary agent which controls the spaceship. Red solid circulars represent bullets fired by the spaceship.}
\end{subfigure}%
\hfill
\begin{subfigure}[t]{0.47\textwidth}
\centering
\includegraphics[height=.7\linewidth]{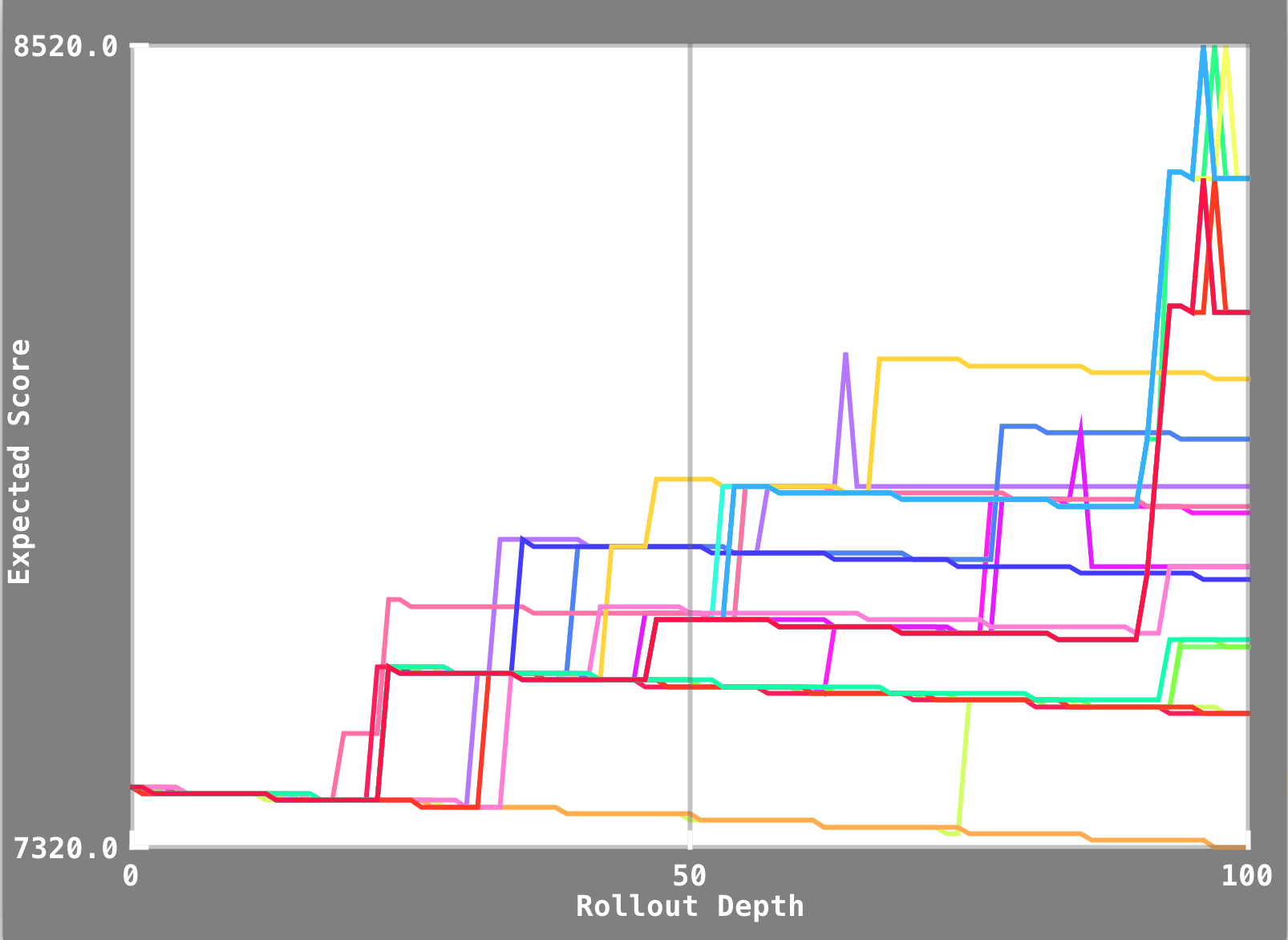}
\caption{\label{fig:ScoreEvolution}Real-time plots showing how the score varies over the planning horizon of the algorithm. The figure shows 10 plots, each of length 100 with a different color. Each of the plots refers to a simulation in Figure \ref{fig:Asteroids}.}
\end{subfigure}%
\caption{
Screenshot of the game screen of Asteroids and real-time plot of game scores in the simulations.}
\end{figure*}

Figure \ref{fig:ScoreEvolution} is an illustration of the plot of game score in real-time by the program to show how score varies over the planning horizon of the algorithm.

\subsection{Solutions Found}

The best solutions found showed common traits for both games,
but with some significant differences.  The best solutions always
used the shift buffer, and set $nResamples$ to 1.  Flipping a bit was normally
preferred, and a large value of $nbMutatedPoints$ of 2.0 or 3.0 gave
best performance.  
The best sequence length for Planet Wars was either
10 or 20, and for Asteroids a sequence length of 100 worked best; the big difference
in performance for various parameter settings demonstrates the importance
and impact of undertaking effective parameter tuning.
The output of the NTBEA provides clear
statistics on the parameter combinations that were tested, but
these run in to pages of output.  We are currently developing
graphical tools to convey this information in a more convenient form.

\section{Conclusion and Future Work}\label{sec:conc}
In this paper, we propose an N-Tuple Bandit Evolutionary Algorithm (NTBEA), combining the strength of N-Tuple systems, multi-armed bandit algorithms and evolutionary algorithms. The proposed approach is compared to two baseline algorithm, Grid Search and Sliding Window compact Genetic Algorithm, on a hyper-parameter optimisation problem: evolving a rolling horizon evolutionary game-playing agent. NTBEA significantly outperformed the baseline algorithms, and
provides an explicit control over exploitation versus exploration.


The next step is the application of the NTBEA to more expensive problems, such as optimising some game-playing agents for playing the 2-player games in the General Video Game AI framework~\cite{gaina2017gvgai}. To ensure the stable performance, the 2-player agent should be evaluated by simulating multiples games against multiple opponent models, which will be computationally expensive.
Additional, it will be challenging to test the NTBEA on larger search space, not only in terms of the dimension number but also the number of legal actions on each of the dimensions.

One thing worth taking into account but not considered in this paper explicitly is the prior knowledge on the dependencies between dimensions. As mentioned previously in Section~\ref{sec:ntbea}, if all the tuples were considered, the number of bandits increases exponentially. Though the use of tuples takes care of the potential dependencies and it is not necessary to consider all the tuples, we still need to decide in advance which are the ones to be included in the model (e.g., all the possible $1$-tuples, $2$-tuples and the single $d$-tuple were included in our test case). Some prior knowledge on the problem can facilitate this selection step and reduces the amount of memory required by NTBEA.

The comparison between the NTBEA and other EDAs will be interesting, and 
more generally a comparison between NTBEA and a wide range of
existing hyper-parameter optimisation algorithms such as SMAC.
It should be noted though that many of the latter algorithms are not set up to 
handle extremely noisy fitness functions such as the games used in this paper.

In summary, the NTBEA is a significant new algorithm that has been shown
to work well both for automated game tuning, and now for optimising a game
playing agent on two very different games.  The algorithm is well suited
to problems where the fitness function is noisy and expensive, gives
good performance, and provides useful statistics on the contribution of each combination of 
parameters as a consequence of the underlying N-Tuple model.


\balance
\bibliographystyle{IEEEtran}
\bibliography{jialinbib}

\begin{thebibliography}{10}
\providecommand{\url}[1]{#1}
\csname url@samestyle\endcsname
\providecommand{\newblock}{\relax}
\providecommand{\bibinfo}[2]{#2}
\providecommand{\BIBentrySTDinterwordspacing}{\spaceskip=0pt\relax}
\providecommand{\BIBentryALTinterwordstretchfactor}{4}
\providecommand{\BIBentryALTinterwordspacing}{\spaceskip=\fontdimen2\font plus
\BIBentryALTinterwordstretchfactor\fontdimen3\font minus
  \fontdimen4\font\relax}
\providecommand{\BIBforeignlanguage}[2]{{%
\expandafter\ifx\csname l@#1\endcsname\relax
\typeout{** WARNING: IEEEtran.bst: No hyphenation pattern has been}%
\typeout{** loaded for the language `#1'. Using the pattern for}%
\typeout{** the default language instead.}%
\else
\language=\csname l@#1\endcsname
\fi
#2}}
\providecommand{\BIBdecl}{\relax}
\BIBdecl

\bibitem{sironi2018self}
C.~F. Sironi, J.~Liu, D.~Perez-Liebana, R.~D. Gaina, I.~Bravi, S.~M. Lucas, and
  M.~H. Winands, ``Self-adaptive mcts for general video game playing,'' in
  \emph{European Conference on the Applications of Evolutionary
  Computation}.\hskip 1em plus 0.5em minus 0.4em\relax Springer, 2018.

\bibitem{kunanusont2017ntuple}
K.~Kunanusont, R.~D. Gaina, J.~Liu, D.~Perez-Liebana, and S.~M. Lucas, ``The
  n-tuple bandit evolutionary algorithm for automatic game improvement,'' in
  \emph{Evolutionary Computation (CEC), 2017 IEEE Congress on}.\hskip 1em plus
  0.5em minus 0.4em\relax IEEE, 2017.

\bibitem{Lehman2011Novelty}
J.~Lehman and K.~O. Stanley, ``Abandoning objectives: Evolution through the
  search for novelty alone,'' \emph{Evolutionary Computation}, no.~2, pp. 198
  -- 223, 2011.

\bibitem{muhlenbein1996recombination}
H.~M{\"u}hlenbein and G.~Paass, ``From recombination of genes to the estimation
  of distributions i. binary parameters,'' in \emph{International conference on
  parallel problem solving from nature}.\hskip 1em plus 0.5em minus 0.4em\relax
  Springer, 1996, pp. 178--187.

\bibitem{larranaga2001estimation}
P.~Larra{\~n}aga and J.~A. Lozano, \emph{Estimation of distribution algorithms:
  A new tool for evolutionary computation}.\hskip 1em plus 0.5em minus
  0.4em\relax Springer Science \& Business Media, 2001, vol.~2.

\bibitem{gonzalez2002mathematical}
C.~Gonz{\'a}lez, J.~A. Lozano, and P.~Larranaga, ``Mathematical modeling of
  discrete estimation of distribution algorithms,'' in \emph{Estimation of
  Distribution Algorithms}.\hskip 1em plus 0.5em minus 0.4em\relax Springer,
  2002, pp. 147--163.

\bibitem{lozano2006towards}
J.~A. Lozano, \emph{Towards a new evolutionary computation: advances on
  estimation of distribution algorithms}.\hskip 1em plus 0.5em minus
  0.4em\relax Springer Science \& Business Media, 2006, vol. 192.

\bibitem{hauschild2011introduction}
M.~Hauschild and M.~Pelikan, ``An introduction and survey of estimation of
  distribution algorithms,'' \emph{Swarm and Evolutionary Computation}, vol.~1,
  no.~3, pp. 111--128, 2011.

\bibitem{decock2013noisy}
J.~Decock and O.~Teytaud, ``Noisy optimization complexity under locality
  assumption,'' in \emph{Proceedings of the twelfth workshop on Foundations of
  genetic algorithms XII}.\hskip 1em plus 0.5em minus 0.4em\relax ACM, 2013,
  pp. 183--190.

\bibitem{rolet2010bandit}
P.~Rolet and O.~Teytaud, ``Bandit-based estimation of distribution algorithms
  for noisy optimization: Rigorous runtime analysis.'' in \emph{LION}.\hskip
  1em plus 0.5em minus 0.4em\relax Springer, 2010, pp. 97--110.

\bibitem{gai2010learning}
Y.~Gai, B.~Krishnamachari, and R.~Jain, ``Learning multiuser channel
  allocations in cognitive radio networks: A combinatorial multi-armed bandit
  formulation,'' in \emph{New Frontiers in Dynamic Spectrum, 2010 IEEE
  Symposium on}.\hskip 1em plus 0.5em minus 0.4em\relax IEEE, 2010, pp. 1--9.

\bibitem{chen2013combinatorial}
W.~Chen, Y.~Wang, and Y.~Yuan, ``Combinatorial multi-armed bandit: General
  framework and applications,'' in \emph{International Conference on Machine
  Learning}, 2013, pp. 151--159.

\bibitem{ontanon2017combinatorial}
S.~Ontan{\'o}n, ``Combinatorial multi-armed bandits for real-time strategy
  games,'' \emph{Journal of Artificial Intelligence Research}, vol.~58, pp.
  665--702, 2017.

\bibitem{NTuple1959}
W.~W. Bledsoe and I.~Browning, ``{Pattern recognition and reading by
  machine},'' in \emph{Proceedings of the Eastern Joint Computer Conference},
  1959, pp. 232 -- 255.

\bibitem{NTupleOthello}
S.~M. Lucas, ``{Learning to Play Othello with N-Tuple Systems},''
  \emph{Australian Journal of Intelligent Information Processing}, vol.~4, pp.
  1--20, 2008.

\bibitem{NTuplePreference}
T.~P. Runarsson and S.~M. Lucas, ``Preference learning for move prediction and
  evaluation function approximation in othello,'' \emph{IEEE Transactions on
  Computational Intelligence and AI in Games}, vol.~6, no.~3, pp. 300--313,
  2014.

\bibitem{liu2017bandit}
J.~Liu, D.~P{\'e}rez-Li{\'e}bana, and S.~M. Lucas, ``Bandit-based random
  mutation hill-climbing,'' in \emph{Evolutionary Computation (CEC), 2017 IEEE
  Congress on}.\hskip 1em plus 0.5em minus 0.4em\relax IEEE, 2017, pp.
  2145--2151.

\bibitem{perez2016general}
D.~Perez-Liebana, S.~Samothrakis, J.~Togelius, S.~M. Lucas, and T.~Schaul,
  ``{General Video Game AI: Competition, Challenges and Opportunities},'' in
  \emph{Thirtieth AAAI Conference on Artificial Intelligence}, 2016, pp.
  4335--4337.

\bibitem{tutorial17cig}
D.~Perez-Liebana, J.~Liu, and S.~M. Lucas, ``General video game {AI} as a tool
  for game design,'' Tutorial at IEEE Conference on Computational Intelligence
  and Games (CIG), 2017.

\bibitem{VisitCountExplorationNIPS2016}
\BIBentryALTinterwordspacing
M.~Bellemare, S.~Srinivasan, G.~Ostrovski, T.~Schaul, D.~Saxton, and R.~Munos,
  ``Unifying count-based exploration and intrinsic motivation,'' in
  \emph{Advances in Neural Information Processing Systems 29}, D.~D. Lee,
  M.~Sugiyama, U.~V. Luxburg, I.~Guyon, and R.~Garnett, Eds.\hskip 1em plus
  0.5em minus 0.4em\relax Curran Associates, Inc., 2016, pp. 1471--1479.
  [Online]. Available:
  \url{http://papers.nips.cc/paper/6383-unifying-count-based-exploration-and-intrinsic-motivation.pdf}
\BIBentrySTDinterwordspacing

\bibitem{bergstra2012random}
J.~Bergstra and Y.~Bengio, ``Random search for hyper-parameter optimization,''
  \emph{Journal of Machine Learning Research}, vol.~13, no. Feb, pp. 281--305,
  2012.

\bibitem{li2016hyperband}
L.~Li, K.~Jamieson, G.~DeSalvo, A.~Rostamizadeh, and A.~Talwalkar, ``Hyperband:
  A novel bandit-based approach to hyperparameter optimization,'' \emph{arXiv
  preprint arXiv:1603.06560}, 2016.

\bibitem{SMAC-LION-2011}
F.~Hutter, H.~H. Hoos, and K.~Leyton-Brown, ``Sequential model-based
  optimization for general algorithm configuration,'' in \emph{Proceedings of
  LION 5}, 2011, pp. 507 -- 523.

\bibitem{bergstra2011algorithms}
J.~S. Bergstra, R.~Bardenet, Y.~Bengio, and B.~K{\'e}gl, ``Algorithms for
  hyper-parameter optimization,'' in \emph{Advances in Neural Information
  Processing Systems}, 2011, pp. 2546--2554.

\bibitem{PopTrainNeuralNets}
\BIBentryALTinterwordspacing
M.~Jaderberg, V.~Dalibard, S.~Osindero, W.~M. Czarnecki, J.~Donahue, A.~Razavi,
  O.~Vinyals, T.~Green, I.~Dunning, K.~Simonyan, C.~Fernando, and
  K.~Kavukcuoglu, ``Population based training of neural networks,''
  \emph{CoRR}, vol. abs/1711.09846, 2017. [Online]. Available:
  \url{http://arxiv.org/abs/1711.09846}
\BIBentrySTDinterwordspacing

\bibitem{rohwer1996theoretical}
R.~Rohwer and M.~Morciniec, ``A theoretical and experimental account of n-tuple
  classifier performance,'' \emph{Neural Computation}, vol.~8, no.~3, pp.
  629--642, 1996.

\bibitem{auer2002finite}
P.~Auer, N.~Cesa-Bianchi, and P.~Fischer, ``Finite-time analysis of the
  multiarmed bandit problem,'' \emph{Machine learning}, vol.~47, no. 2-3, pp.
  235--256, 2002.

\bibitem{liu2016rolling}
J.~Liu, D.~P{\'e}rez-Li{\'e}bana, and S.~M. Lucas, ``Rolling horizon
  coevolutionary planning for two-player video games,'' in \emph{Computer
  Science and Electronic Engineering (CEEC), 2016 8th}.\hskip 1em plus 0.5em
  minus 0.4em\relax IEEE, 2016, pp. 174--179.

\bibitem{gaina2017analysis}
R.~D. Gaina, J.~Liu, S.~M. Lucas, and D.~P{\'e}rez-Li{\'e}bana, ``Analysis of
  vanilla rolling horizon evolution parameters in general video game playing,''
  in \emph{European Conference on the Applications of Evolutionary
  Computation}.\hskip 1em plus 0.5em minus 0.4em\relax Springer, 2017, pp.
  418--434.

\bibitem{gaina2017population}
R.~D. Gaina, S.~M. Lucas, and D.~P{\'e}rez-Li{\'e}bana, ``Population seeding
  techniques for rolling horizon evolution in general video game playing,'' in
  \emph{Evolutionary Computation (CEC), 2017 IEEE Congress on}.\hskip 1em plus
  0.5em minus 0.4em\relax IEEE, 2017, pp. 1956--1963.

\bibitem{gaina2017rolling}
R.~D. Gaina, S.~M. Lucas, and D.~Perez-Liebana, ``Rolling horizon evolution
  enhancements in general video game playing,'' in \emph{Computational
  Intelligence and Games (CIG), 2017 IEEE Conference on}.\hskip 1em plus 0.5em
  minus 0.4em\relax IEEE, 2017, pp. 88--95.

\bibitem{liu2017evolving}
J.~Liu, J.~Togelius, D.~P{\'e}rez-Li{\'e}bana, and S.~M. Lucas, ``Evolving game
  skill-depth using general video game ai agents,'' in \emph{Evolutionary
  Computation (CEC), 2017 IEEE Congress on}.\hskip 1em plus 0.5em minus
  0.4em\relax IEEE, 2017, pp. 2299--2307.

\bibitem{lucas2017efficient}
S.~M. Lucas, J.~Liu, and D.~P{\'e}rez-Li{\'e}bana, ``Efficient noisy
  optimisation with the multi-sample and sliding window compact genetic
  algorithms,'' in \emph{Computational Intelligence (SSCI), 2017 IEEE Symposium
  Series on}.\hskip 1em plus 0.5em minus 0.4em\relax IEEE, 2017.

\bibitem{gaina2017gvgai}
R.~D. Gaina, A.~Couetoux, D.~J. N.~J. Soemers, M.~H.~M. Winands, T.~Vodopivec,
  F.~Kirchge$\beta$ner, J.~Liu, S.~M. Lucas, and D.~Perez-Liebana, ``The 2016
  two-player {GVGAI} competition,'' \emph{IEEE Transactions on Computational
  Intelligence and AI in Games}, 2017.

\end{thebibliography}

\end{document}